\documentclass[letterpaper]{article}
\usepackage{aaai}
\usepackage{graphicx}
\usepackage{times}
\usepackage{helvet}
\usepackage{courier}
\usepackage{verbatim}
\begin{document}
%
\title{Visual Sentiment Prediction with Deep Convolutional Neural Networks }

\author{Can Xu\thanks{Part of this work was completed while the first author was interning at Yahoo Labs.} \\
University of California, San Diego\\
canxu@ucsd.edu\\
\And
 Suleyman Cetintas, Kuang-Chih Lee, Li-Jia Li\\
 Yahoo Labs, Sunnyvale, CA\\
\{cetintas, kclee, lijiali\}@yahoo-inc.com\\
}

\maketitle
\begin{abstract}
\begin{quote}
Images have become one of the most popular types of media through which users convey their emotions within online social networks. Although vast amount of research is devoted to sentiment analysis of textual data, there has been very limited work that focuses on analyzing sentiment of image data. In this work, we propose a novel visual sentiment prediction framework that performs image understanding with Convolutional Neural Networks (CNN). Specifically, the proposed sentiment prediction framework performs transfer learning from a CNN with millions of parameters, which is pre-trained on large-scale data for object recognition. Experiments conducted on two real-world datasets from Twitter and Tumblr demonstrate the effectiveness of the proposed visual sentiment analysis framework. 
\end{quote}
\end{abstract}

\section{Introduction}
Writing and sharing posts have become one of the most popular activities in major social network services for communication and information exchange in the world. Opinions and emotions are important concepts embedded in posts to show friendship and social support. Algorithms to identify sentiment can be helpful to understand such user behaviors and therefore are widely applicable to many applications, such as blog recommendation, behavior targeting, and viral marketing. Many existing research papers \cite{textsenti,sentistrength} have focused on opinion mining and sentiment analysis of text information in the post. 

\begin{figure}[t]
\centering
\includegraphics[width=0.48\textwidth]{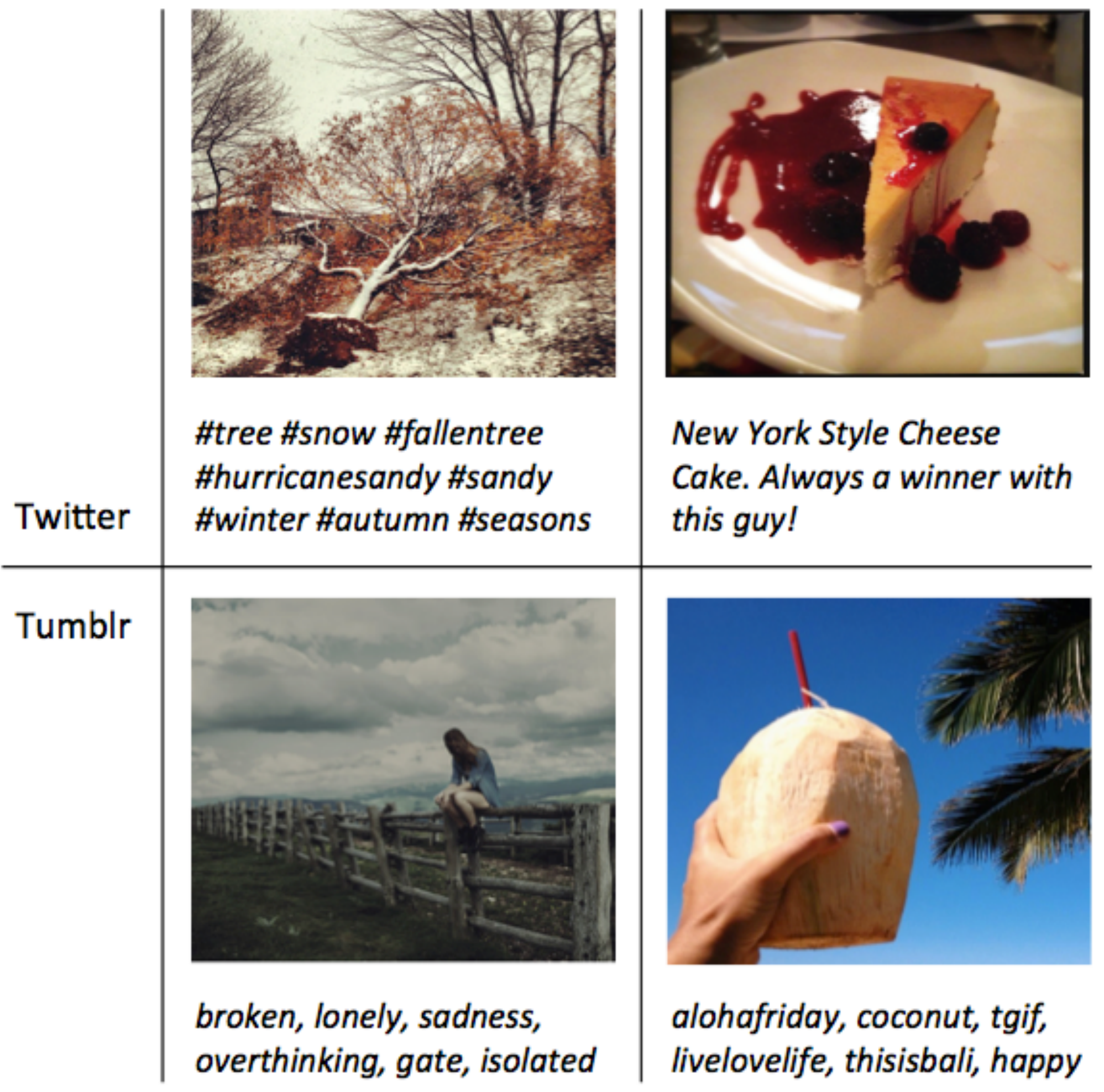}
\caption{Examples of negative and positive images and associated tags/text from Twitter and Tumblr. }
\label{fig:example}
\end{figure}

However, according to existing surveys \cite{Sentribute,Chang:2014}, multimedia contents become more popular in social networks especially on platforms such as Tumblr, Instagram, and Flickr. In addition, a big majority of those photo/video posts contain only small amount of short tags or do not contain any text at all. Therefore, lots of opinions and emotions are conveyed by visual contents alone. For example, Figure~\ref{fig:example} shows two image posts on Tumblr, where opposite emotions are obviously characterized through visual cues and contexts. However, most of the state-of-the-art image understanding algorithms in computer vision are designed for the problems of object recognition and scene classification, leaving the emotional aspects of images relatively unexplored. 

In this paper, we focus on the problem of sentiment prediction purely based on the visual information within a blog post. Instead of focusing on defining and training mid-level attributes related to emotional perception \cite{Sentribute,Borth:Sentiment}, we propose a novel framework that efficiently transfers CNNs learned on a large-scale dataset to the task of visual sentiment prediction. Our transfer learning has a major advantage over those standard approaches because there is no requirement of domain knowledge from psychology or linguistics, and in consequence, the simplicity of the training process makes the framework be deployed and scaled easily in the production environment. We show that the transferred network activations consistently outperform the state-of-the-art methods by a large margin. In addition to the existing benchmarks that only includes positive or negative labels for images \cite{Borth:Sentiment}, we also  introduce a 5-scale sentiment rating, which accounts for neutral images and different sentiment strength of the same polarity. We construct a dataset from Tumblr images annotated with those fine-grained sentiment scores. The effectiveness of the labeling scheme is analyzed and validated from the statistics of multiple annotators.

The rest of the paper is organized as follows. In Section \ref{sec:relatedwork}, we review some closely related work. Next we present our proposed sentiment prediction framework in Section \ref{sec:method}. Then we take a closer look at the Tumblr and Twitter datasets and study some of its important characteristics in Section \ref{sec:dataset}. We detail our experimental methodology in Section \ref{sec:experimentalmethodology}. Experimental results are shown in Section \ref{sec:expresults} followed by conclusions in Section \ref{sec:conclusions}.

\section{Related Work}
\label{sec:relatedwork}
While research on sentiment prediction of visual content is far behind, extensive research has been conducted on opinion mining and sentiment analysis of text, and a comprehensive survey can be found in \cite{textsenti}. Previous work on visual sentiment analysis has mostly been conducted to develop mid-level attributes for selecting features from low-level image features. \cite{Sentribute} generated mid-level attributes from scene and facial expression dataset to describe the visual phenomena in a scene perspective as well as incorporating facial emotion detectors when faces are present in the image. \cite{Borth:Sentiment} built large-scale Visual Sentiment Ontology based on psychological theories and web mining and trained detectors of selected visual concepts for sentiment analysis. ~\cite{Video} evaluated the performance of different low-level descriptors and mid-level attributes as visual features for sentiment classification and showed that semantic-level clues are effective for predicting emotions. The major drawback for those approaches is that the training process requires lots of domain knowledge of psychology or linguistics to define the mid-level attributes, and human intervention to fine tune the sentiment prediction results.   

Though hand-engineered image descriptors such as color histogram, HOG~\cite{HOG}, SIFT\cite{SIFT} etc. have been shown effective in object recognition and image classification, deep compositional architectures\cite{Alex} have recently outperformed all known image classification pipelines on ImageNet large scale visual recognition challenge (ILSVRC) 2012 \cite{ImageNet}.  Deep convolutional neural networks (CNN)\cite{Alex} are layered classifiers with millions of parameters. With the advent of large-scale labeled data\cite{ImageNet}, fully-supervised CNNs are able to learn a deep representation without overfitting the huge amount of the parameters. 

The estimation of CNN parameters requires a very large amount of annotated data. There has been extensive work that perform transfer learning across different domains. \cite{le} reported success with transferring deep representations to small datasets as CIFAR and MINST. Recent studies \cite{Decaf} \cite{Transfer} show that the parameters of CNN trained on large-scale dataset such as ILSVRC can be transferred to object recognition and scene classification tasks when the data is limited, resulting better performance than traditional hand-engineered representations. Our work is motivated by \cite{Transfer} a lot, and we apply the concept of transfer learning Deep CNN from large-scale image classification to the problem of sentiment prediction. 

\section{Method}
\label{sec:method}

\begin{figure*}[t]
\centering
\includegraphics[width=1\textwidth]{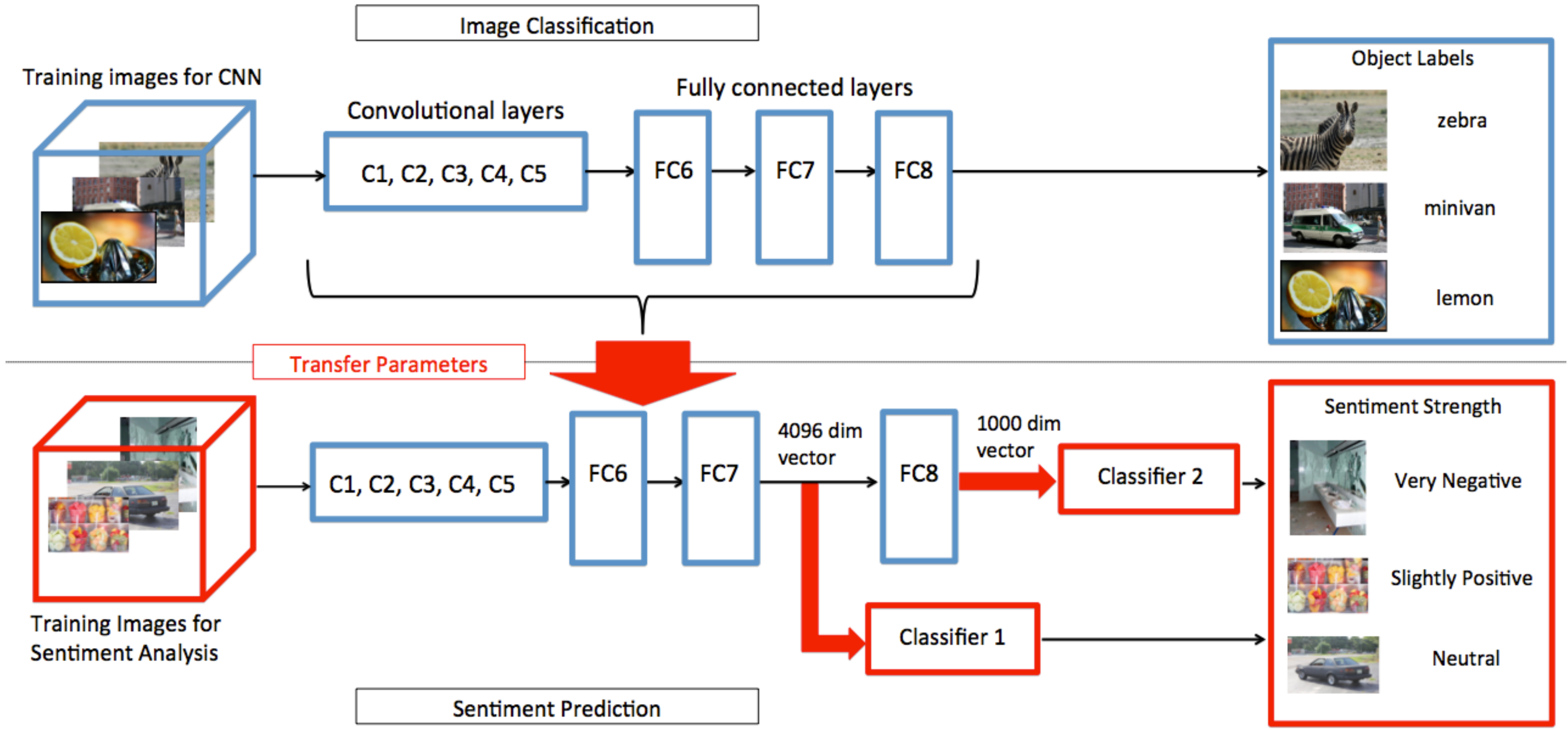}
\caption{The overall architecture of the proposed visual sentiment prediction framework. The CNN is first trained on a large-scale dataset (ImageNet) for image classification. The parameters of five convolutional layers (C1 to C5) and three fully connected layers (FC6 to FC8) are then transferred to the problem of sentiment prediction for generating image representations. Two types of activations are used as image-level features, namely the 4096-dimension output from fc7 and the 1000-dimension output from fc8; and two classifiers are trained with the two sets of extracted features on the sentiment dataset respectively.}
\label{fig:pipeline}
\end{figure*}

This section introduces a comprehensive computational framework for visual sentiment prediction via deep image understanding by utilizing Convolutional Neural Networks (CNN). Details of the overall architecture of the proposed framework can be seen in Figure~\ref{fig:pipeline}. The network is first trained on a large-scale image dataset for object classification, and then the learned parameters of the network are transferred to the task of sentiment prediction for generating image-level representations. Finally classifiers are trained from the features extracted from the sentiment images. 

In the following subsections, we first introduce the deep learning model \cite{Alex} that has achieved a leap in image classification recently. Next, we discuss the activations generated from the pre-trained CNN and consider the output from certain layers of CNN as image-level representation for the new task of sentiment prediction.  

\subsection{Deep convolutional neural networks}
For the pre-trained CNN, we use the deep architecture of \cite{Alex}. The CNN is composed of seven internal layers and ultimately a soft-max layer. The hidden layers are five successive convolutional layers followed by two fully connected layers. The nonlinearity of each neuron in this CNN is modeled by Rectified Linear Units (ReLUs) $f(x)=\max(0,x)$, which accelerates learning compared with saturating  nonlinearity such as $\tanh$ units. The CNN takes a 224 $\times$ 224 pixel RGB image as input. Each convolutional layer convolves the output of its previous layer with a set of learned kernels, followed by ReLU non-linearity, and two optional layers, local response normalization and max pooling. The local response normalization layer is applied across feature channels, and the max pooling layer is applied over neighboring neurons. The fifth convolutional layer is followed by two fully connected layers each of which has 4096 neurons. The output of the 7th layer is fed into the last soft-max layer, which produces a distribution over the pre-defined classes.

\subsection{Network training}

We use the open source implementation named Caffe \cite{Caffe}, which implements the network
of \cite{Alex} to train the CNN on ILSVRC-2012 dataset. It is a subset of ImageNet, consisting of around 1.2 million labeled data with 1000 different classes. All the images in ILSVRC-2012 are quality-controlled and human-annotated for the presence or absence of 1000 object categories. The network is trained to maximize the multinomial logistic regression
objective via back propagation using stochastic gradient descend.

\subsection{Transfer learning}
As discussed above, the deep CNN is first trained in a fully-supervised way with large-scale annotated data for the image classification task. Then the learned parameters are transferred to the task of sentiment prediction, where the images are from a different domain and the labeled data is limited. We extract two types of image-level representations from the network, which will be discussed in the following.

In the first setting, we remove the soft-max layer while keep all the parameters in the internal layers of the pre-trained CNN fixed. We consider the activations from the 7th layer neurons as the image-level representation, which is a 4096 dimension feature. As is demonstrated by \cite{Decaf}, the 7th layer output of pre-trained CNN generalizes well to object recognition and detection. In this setting, we explore its capacity to higher-level concept understanding, namely sentiment.

In the other setting, we keep all the parameters from the network, including the 1000-way classifier in the soft-max layer. The image-level representation is the 1000 dimension vector of the distribution over the object categories of ILSVC. In other tasks such as object detection or subcategory recognition \cite{Decaf}, only the activations from the 7th layer or previous layers are considered as image representations. For the problem of sentiment prediction, we consider this 1000D distribution score as another high level attribute descriptor, which is directly associated with the objects that appear in the image. 

\subsection{Classification}
With the image-level representation, sentiment prediction models can be easily trained with linear classifiers. In \cite{Borth:Sentiment}, it is shown that Logistic Regression model leads to better performance than SVM classifiers with sentiment features. In this work, we employ Logistic Regression as the classifier on top of the generated features. For the 4096D activations and 1000D soft-max response, a Logistic regression model is trained with each type of features.

\section{Dataset}
\label{sec:dataset}

\begin{table}[t]
\centering
\caption{Statistics of the Tumblr dataset. Each column presents the number of images in each of the 5-level sentiment labels.}
\vspace{3mm}
\begin{tabular}{p{1in} p{0.2in} p{0.2in} p{0.2in} p{0.2in} p{0.2in}} \hline \hline
 & \multicolumn{5}{ c }{Sentiment Label} \\ 
  & -2 & -1 & 0 &  1 &  2 \\ \hline
\# of Images  & 165 & 190 & 90 & 465 & 200 \\ \hline\hline
\end{tabular}
\label{tab:stats}
\end{table}

\begin{figure*}[t]
\centering
\includegraphics[width=1\textwidth]{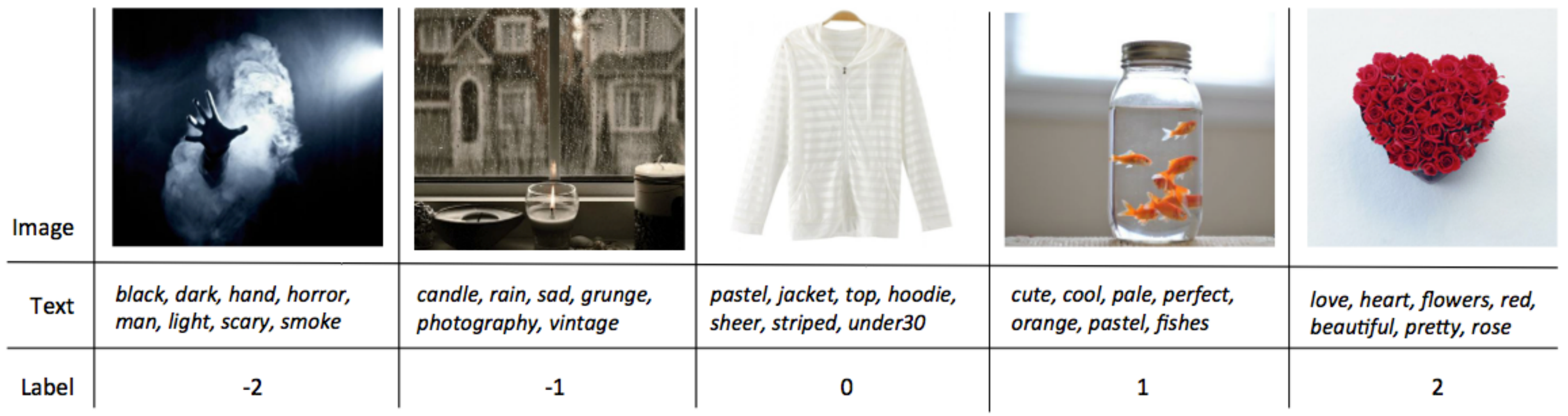}
\caption{Example images and text/tags together with the corresponding sentiment labels from the Tumblr dataset. Each column is the visual and text data together with the corresponding ground truth for sentiment strength. The first row lists the images; the second row lists the tags/text corresponding to the image; and the third row presents the sentiment score with image-text combined inspection.}
\label{fig:finegrain}
\end{figure*}

We evaluate the proposed methods on two real-world datasets from two major microblogging sites, namely Twitter and Tumblr. The Twitter dataset is a public dataset and has been used in prior work \cite{Borth:Sentiment}. Details of the Twitter dataset can be found in Section~\ref{sec:dataset:Twitter}. The Tumblr dataset is a proprietary dataset that we collect for this work, and the details of the data collection and ground truth labeling are discussed in Section~\ref{sec:dataset:Tumblr}.


\subsection{Twitter}
\label{sec:dataset:Twitter}
The Twitter benchmark \cite{Borth:Sentiment} is collected from image tweets and labeled via combined image-text for sentiment polarity, i.e, the annotators are given both the image and the text associated with the image. It includes 470 positive tweets and 133 negative tweets. 

\subsection{Tumblr}
\label{sec:dataset:Tumblr}
Although Tumblr is ranked as the 2nd largest microblogging service after Twitter, there has been very limited research on content analysis of Tumblr\footnote{The reported datasets and results are deliberately incomplete and subject to anonymization, and thus do not necessarily reflect the real portfolio at any particular time.}. In this work, we provide, to the best of our knowledge, the first study on the analysis of the photo posts on Tumblr. As is discussed in the Introduction section, Tumblr provides a rich repository of images and tags that are associated with users' sentiment. We construct a visual sentiment dataset from the photo posts on Tumblr. The details for data collection and ground truth labeling are discussed in the following subsections. 


\subsubsection{Data collection}

We utilize the tags/hashtags in the Tumblr photo posts to pre-select images that have detectable sentiment content. Typically, the tags indicate the users' sentiment for the uploaded images. We apply a subjectivity clue lexicon \cite{lexicon} to the sampling process. This lexicon list the neutrality or polarity of 8222 frequently used words, with subjectivity clues out of five sentiment levels, namely \textit{strong negative, weak negative, neutral, weak positive and  strong positive}. For instance, according to the lexicon, \textit{happy} is positive in the strongly subjective sense while \textit{sad} is negative in the strongly subjective sense. We use the lexicon to identify sentiments from the word level, and only consider the images with at least one tag that has polarity in the strongly subjective sense in the sampling process. Specifically from a large-pool of photo posts on Tumblr, we randomly sampled and collected 1179 photo posts for human labeling with the sentiment lexicon applied to words in the tags. Note that, although the textual information provides strong prior emotional clues for the corresponding image, they are far from effective for recognizing the sentiment of the post since the short text in tags suffers from lack of context. For instance, "mar " in the lexicon is negative for that it means impair when used as verb, while it is also short for March and tagged by users to indicate the month. In \cite{Borth:Sentiment}, experiments show that text based methods have lower prediction accuracy than visual based method on data with short text such as tweets.

%


\subsubsection{Ground truth labeling} 

To obtain ground truth of the collected photo posts, we asked 5 annotators (2 females and 3 males) to label the data. Each image as well as the associated tags were assigned to exactly 3 annotators. Annotators are asked to provide a sentiment score out of a 5-scale labeling scheme ranging from -2 to 2, namely \textit{strongly negative, weakly negative, neutral, weakly positive and strongly positive}. Although this is different than using a simpler bi-polar labeling scheme as has been used in the Twitter dataset introduced in the previous Section by a prior work \cite{Borth:Sentiment}, we think that capturing sentiment strength is also important along with capturing sentiment polarity. Indeed, fine-grained categorization in sentiment strength is widely accepted in text analysis \cite{sentistrength}. Fine-grained ratings (e.g., with 5 degrees of scale) have also been commonly used for quantifying the opinions of raters and/or labelers in many real-world applications such as large-scale movie ratings in recommender systems \cite{netflix}, crowdsourcing experiments to quantify similarity between professionals in social networks \cite{Cetintas2011} and between folk narratives in literature \cite{Nguyen2014}, etc.

\begin{table*}[th]
\centering
\caption{Results of the proposed fc7 and fc8 methods on the benchmark Twitter dataset in comparison to the Low-level Features ~\cite{Borth:Sentiment} and SentiBank ~\cite{Borth:Sentiment} baselines. fc7 denotes the 4096D feature from the 7th fully connected layer and fc8 denotes the 1000D classification score from the 8th soft-max layer. The performance is evaluated by the AUC.}
\vspace{3mm}
\begin{tabular}{p{2.5in} p{0.6in} p{0.6in} p{0.6in}} \hline \hline
& \multicolumn{3}{ c }{Sentiment Label} \\
& Positive & Negative & Overall \\ \hline
Low-level Features ~\cite{Borth:Sentiment} & 0.500 & 0.516 & 0.508\\ 
SentiBank ~\cite{Borth:Sentiment} & 0.516 & 0.511 & 0.514\\ 
fc7 (Proposed method) & \textbf{0.648} & \textbf{0.649} & \textbf{0.649}\\ 
fc8 (Proposed method) & 0.619 &  0.610 & 0.615\\ \hline \hline
\end{tabular}
\label{tab:tweet}
\end{table*}

After collecting all the annotations, we took the majority vote out of the 3 scores for each image; that is, an image is considered valid only when at least 2 of the 3 annotators agree on the exact label (out of 5 possible labels). Overall, a set of 1110 images is collected with image-text combined ground truth out of the total 1179 images. This corresponds to an sentiment label agreement percentage of 94\%, which is indeed very high given the 5-level granularity. This explicitly shows that utilizing the fine-grained, 5-level sentiment strength is a better choice for describing sentiment strength in visual content.

\section{Experiments}
\label{sec:experimentalmethodology}

In this section, we describe two baseline methods, namely low-level visual features and SentiBank \cite{Borth:Sentiment}, in comparison with our proposed approaches. Then we introduce the evaluation metric used in the performance evaluation. Finally, we present the experimental results of the proposed approaches as well as the baselines.

\subsection{Baselines}
\label{sec:baselines}
\subsubsection{Low-level visual features}
It has recently been shown that a set of low-level visual features can be useful for characterizing sentiment clues such as scenes, textures, faces as well as other abstract concepts \cite{Borth:Sentiment}. Therefore, we follow the same setup in \cite{Borth:Sentiment} and extract a set of generic low-level visual features as the first baseline in this work. Specifically, we extract features including a 3 $\times$  256 dimension RGB Color Histogram, a 512 dimension GIST\cite{gist} descriptor, a 53 dimension LBP descriptor and a Bag-of-Words descriptor using a 1,000 word dictionary with max pooling over a 2-layer spatial pyramid.

\subsubsection{SentiBank}
SentiBank, first introduced in \cite{Borth:Sentiment}, is a new concept representation that includes 1200 concepts.  Each of the concept is defined as an adjective-noun pair, e.g. colorful clouds, crying baby, misty night, etc. The ontology is constructed based on psychology studies and web mining, while the detector for each concept is trained on Flickr images. We use the SentiBank approach as the second baseline in this work.

\subsection{Evaluation Metric}
In all the experiments, we use Area Under the receiver operating characteristic Curve (AUC) as the metric for performance evaluation. AUC is a widely used metric for classification because it describes the discriminating power of the classifier in general, and is independent of different decision criteria \cite{AUCTutorial}. Note that the previous work \cite{Borth:Sentiment} performed the sentiment prediction on the imbalanced Twitter dataset using the metric of prediction accuracy, and the results for their visual and text+visual based methods are 0.70 and 0.72, respectively. However, a naive classifier can achieve a better prediction accuracy of 0.78 by simply making all the decisions towards to the majority class, i.e., the positive sentiment in this case. In contrast, AUC is less sensitive to imbalanced datasets \cite{imbalance}; and therefore, it is a more appropriate evaluation measure to evaluate the performance for the sentiment prediction task, since the majority of uploaded images in social networks typically belong to the positive sentiment polarity class. 

Following the setup in ~\cite{Borth:Sentiment}, all results are calculated based on the average AUC of five independent runs with five partitions of the dataset.

\subsection{Experimental Results}
\label{sec:expresults}

\begin{table*}[t]
\centering
\caption{ Results of the proposed fc7 and fc8 methods on the Tumblr dataset in comparison to the Low-level Features~\cite{Borth:Sentiment} and SentiBank~\cite{Borth:Sentiment} baselines. fc7 denotes the 4096D feature from the 7th fully connected layer and fc8 denotes the 1000D classification score from the 8th soft-max layer. The performance is evaluated by the AUC. }
\vspace{3mm}
\begin{tabular}{{p{2.5in} p{0.5in} p{0.5in} p{0.5in} p{0.5in} p{0.5in} p{0.5in}}} \hline \hline
& \multicolumn{6}{ c }{Sentiment Label} \\ 
  & -2 & -1 & 0 & 1 & 2 & Overall \\ \hline
Low-level Features ~\cite{Borth:Sentiment} 	& 0.716 & 0.664 & 0.602 & 0.601 & 0.655 & 0.646 \\ 
SentiBank ~\cite{Borth:Sentiment} 			& 0.745 & 0.684 & 0.635 & 0.638 & 0.686 & 0.677 \\ 
fc7 (Proposed method) 		& \textbf{0.801}	& 0.677 & \textbf{0.692} & \textbf{0.673} & 0.694 & \textbf{0.704} \\ 
fc8 (Proposed method) 		& 0.783 & 0.679 & \textbf{0.680} & \textbf{0.682} & 0.683 & \textbf{0.701} \\ \hline \hline
\end{tabular}
\label{tab:tumblr}
\end{table*}

This subsection presents the experimental results of the proposed approaches presented in Section \ref{sec:method} as well as the baseline approaches presented in Section \ref{sec:baselines} for the visual sentiment detection task. All approaches are evaluated on the datasets presented in Section \ref{sec:dataset}.

First, we compare the performances of the baseline approaches of Low-level features and SentiBank with respect to each other. It can be seen in Tables \ref{tab:tweet} and \ref{tab:tumblr} that SentiBank outperforms Low-level features approach on both Twitter and Tumblr datasets. This shows the power of the SentiBank approach in being to better capture emotional concepts that better reflect the sentiments in images. It is also important to note that the performance difference on the Tumblr dataset is bigger than the difference on Twitter dataset. We attribute this to the fact that the Twitter dataset is a much more noisier dataset than the Tumblr dataset, and SentiBank approach is able to benefit more from a cleaner dataset.

Second, we compare the performances of the proposed fc7 and fc8 methods in comparison to the SentiBank and Low-level features baseline approaches. It can be seen in Tables \ref{tab:tweet} and \ref{tab:tumblr} that both fc7 and fc8 approaches outperform both baselines on both Twitter and Tumblr datasets. Specifically, fc7 and fc8 outperform the baselines approaches with a very large margin on the Twitter dataset compared to the Tumblr dataset. This observation can be explained by the previous observation that SentiBank approach suffers the noisy data in Twitter dataset, and is not able to reach its full potential. Yet, on the Tumblr dataset, it can be seen that SentiBank outperforms the Low-level features approach, and achieves closer (although still significantly worse) results than the proposed fc7 and fc8 approaches. This set of results clearly demonstrates that the proposed visual sentiment prediction framework is able to successfully utilize the power of the pre-trained Convolutional Neural Network by transferring domain knowledge from the image classification domain to the sentiment prediction domain, and by effectively utilizing the fc7 and fc8 representations of the images in its sentiment prediction classifiers. Therefore, the results provided in this section, for the first time, suggest that Convolutional Neural Networks are highly promising for visual sentiment analysis.

Next, we compare the proposed fc7 and fc8 approaches in comparison to each other. It can be seen in Tables \ref{tab:tweet} and \ref{tab:tumblr} that the proposed fc7 approach outperforms the proposed fc8 approach on Twitter dataset, and achieves comparable results with the proposed fc8 approach on the Tumblr dataset. This can be explained by the fact that the fc7 representation demonstrates superior performance compared with fc8 since the activations from the 7th layer of CNN characterizes more aspects of the image than object detection scores in 8th layer. Yet, on the Tumblr dataset, the performances of the fc7 and fc8 approaches are comparable. This can also be explained by the fact that Tumblr dataset is a much cleaner dataset than the Twitter dataset, and it is relatively easier for the fc8 approach to find objects in the images of the Tumblr dataset that it can associate with sentiments. This set of results suggests that fc7 and fc8 approaches can be used interchangeably when the sentiment dataset of interest is clean and has images that have relatively easier objects to be identified.

\section{Conclusions}
\label{sec:conclusions}
Sentiment analysis is an important task for ads and recommendation. While vast majority of previous works of sentiment analysis on social web were conducted on text, we propose to focus on the analysis of images, one of the dominant media types of online microblogging services. In this paper, a novel sentiment analysis framework based upon convolutional neural network is introduced for visual sentiment prediction. We show that the image representations from the CNN trained on a large-scale dataset could be efficiently transferred for sentiment analysis. To evaluate the proposed method on real-world data, we constructed a sentiment benchmark from the photo posts on Tumblr, which has a rich repository of images and associated tags reflecting users' emotions. We also introduce a 5-scale granularity of sentiment rating, which is more comprehensive compared with the bi-polar labeling scheme in the existing datasets. Experiments on existing Twitter dataset \cite{Borth:Sentiment} demonstrate that our proposed models outperform the state-of-the-art methods on both Twitter and Tumblr datasets.

There are several interesting future directions for us to explore. First, we intend to adapt the CNN to the sentiment images with the user-tagged data of Tumblr via semi-supervised learning. Given the vast amount of images and the associated tags posted by users, we will do domain-specific fine-tuning to the parameters of the fully connected layers in the CNN. Furthermore, we would like to apply our research results to many applications in different domains, such as recommendation, advertising, and games. Finally, we would like to extend our sentiment research to video data as well. 

\bibliographystyle{aaai}
\bibliography{VisualSentPred_with_DCNN}  

\end{document}